\documentclass[journal]{IEEEtran}
\ifCLASSOPTIONcompsoc
  \usepackage[nocompress]{cite}
\else
  \usepackage{cite}
\fi
\ifCLASSINFOpdf
\else
\fi
\usepackage{graphicx}
\usepackage{cite}
\usepackage{bm}
\usepackage{booktabs}
\usepackage{amsmath, amsfonts, amssymb}    
\usepackage{algorithm}
\usepackage{algpseudocode}
\DeclareMathOperator*{\argmax}{arg\,max}

\usepackage{hyperref}
\usepackage{booktabs}
\usepackage{multirow}
\usepackage{caption}
 \usepackage[caption=false,font=footnotesize]{subfig}
\usepackage{float}
\usepackage{tabularx} 
\newfloat{algorithm}{t}{lop}
\newcolumntype{Y}{>{\centering\arraybackslash}X}
\usepackage{pifont}
\hypersetup{
 linkcolor=red,
 citecolor=blue,
 colorlinks=true
}

\begin{document}
%
\title{Trajectory Forecasts in Unknown Environments Conditioned on Grid-Based Plans}
%
%
%
%

\author{Nachiket~Deo,
       and~Mohan~M.~Trivedi,~\IEEEmembership{Fellow,~IEEE}
\thanks{The authors are with the Laboratory for Intelligent and Safe Automobiles
(LISA), University of California at San Diego, La Jolla, CA 92093 USA}
\thanks{(email:ndeo@ucsd.edu, mtrivedi@ucsd.edu)}
\thanks{}}
\maketitle

\begin{abstract}
We address the problem of forecasting pedestrian and vehicle trajectories in unknown environments, conditioned on their past motion and scene structure.
Trajectory forecasting is a challenging problem due to the large variation in scene structure and the multimodal distribution of future trajectories. Unlike prior approaches that directly learn one-to-many mappings from observed context to multiple future trajectories, we propose to condition trajectory forecasts on \textit{plans} sampled from a grid based policy learned using maximum entropy inverse reinforcement learning (MaxEnt IRL). We reformulate MaxEnt IRL to allow the policy to jointly infer plausible agent goals, and paths to those goals on a coarse 2-D grid defined over the scene. We propose an attention based trajectory generator that generates continuous valued future trajectories conditioned on state sequences sampled from the MaxEnt policy. Quantitative and qualitative evaluation on the publicly available Stanford drone and NuScenes datasets shows that our model generates trajectories that are \textit{diverse}, representing the multimodal predictive distribution, and \textit{precise}, conforming to the underlying scene structure over long prediction horizons. 
\end{abstract}


\IEEEdisplaynontitleabstractindextext

%
\IEEEpeerreviewmaketitle

\section{Introduction}
\label{sec:introduction}


Autonomous vehicles need to operate in a space shared with humans and human driven vehicles. In order to plan safe and efficient paths through complex traffic, autonomous vehicles need the ability to reason about the intent and future motion of surrounding agents. We address the problem of predicting the future locations of pedestrians and vehicles, conditioned on their track history and a bird's eye view representation of the static scene around them. In particular, we wish to forecast trajectories in \textit{unknown environments}, where prior observations of trajectories are unavailable. This is a challenging task due to a number of factors:
 
\begin{figure*}[t]
\centering
\includegraphics[width=\linewidth]{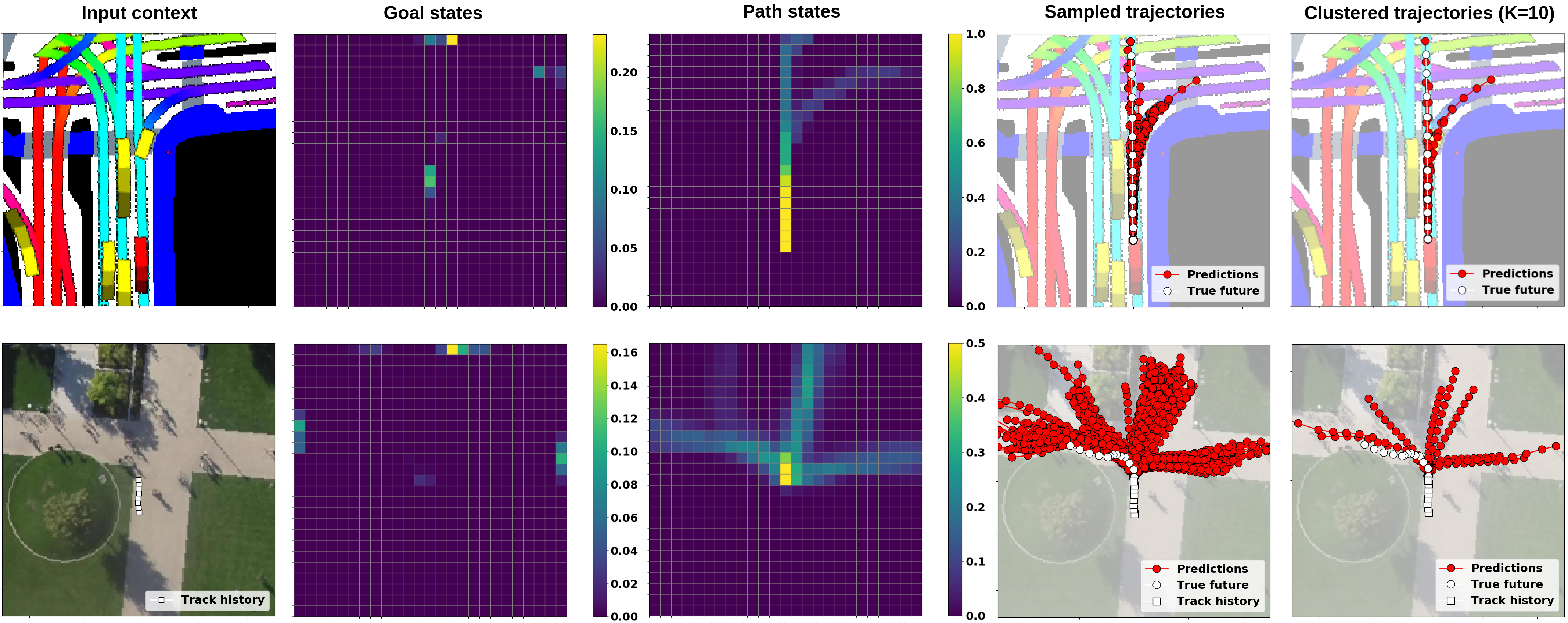}
\caption{\textbf{Forecasts generated by P2T:}  We address the problem of forecasting agent trajectories in unknown environments. The inputs to our model \textit{(left)} are snippets of the agents' past trajectories, and a bird's eye view representation of the scene around them. Our model infers potential goals of the agents \textit{(left-middle)} and paths to these goals \textit{(middle)} over a coarse 2-D grid defined over the scene by modeling the agent as a MaxEnt policy exploring the grid. It generates continuous valued trajectories conditioned on the grid-based plans sampled from the policy \textit{(middle-right)}. Finally it outputs \textit{K} predicted trajectories by clustering the sampled trajectories \textit{(right)}.}
\label{fig_1}
\end{figure*}

\begin{itemize}
    \item \textbf{Unknown goals and path preferences:} Without prior observations of agent trajectories in a scene, goals and path preferences of agents need to be inferred purely from the scene layout.
    \item \textbf{Scene-compliance:} Additionally, the predicted trajectories need to conform to the inferred goals and paths in the scene layout.
    \item \textbf{Variability in scene structure:} Scene elements such as roads, sidewalks, crosswalks and buildings can be found in a variety of configurations. Thus there's high variability in the inputs to the trajectory forecasting model.
    \item \textbf{Non-linearity of agent trajectories:} Drivers and pedestrians can make several decisions over long prediction horizons, leading to highly non-linear trajectories. Thus, there's high variability in the outputs of the trajectory forecasting model. 
    \item \textbf{Multimodality:} Finally, the distribution of future trajectories is multimodal. In any given scene, an agent can have one of multiple potential goals, with multiple paths to each goal. Regression based approaches have been shown to average the modes of the predictive distribution \cite{lee2017desire, gupta2018social, deo2018convolutional}. This would lead to trajectory forecasts that may not conform to the underlying scene. 
\end{itemize}


Recent work has addressed multimodality of the distribution of future trajectories by learning one-to-many mappings from input context to multiple trajectories. Mixture models \cite{deo2018multi, cui2019multimodal, deo2018convolutional, zyner2019naturalistic, casas2018intentnet, ridel2019scene, messaoud2020multi,hong2019rules} assign a mixture component to each mode of the trajectory distribution. They output mean trajectories and probabilities for each mixture component, along with prediction uncertainty. Alternatively, conditional generative models \cite{gupta2018social, sadeghian2018sophie, zhao2019multi, lee2017desire, rhinehart2018r2p2, bhattacharyya2019conditional, bhattacharyya2018accurate, rhinehart2019precog} map input context and a sample from a simple latent distribution to a trajectory output. They can be sampled from indefinitely, to output multiple trajectories. Both, conditional generative models and mixture models need to learn a mapping from a high dimensional input space (variable scene and agent configuration) to a high dimensional output space (continuous valued trajectories). Several recent works thus incorporate inductive bias into the predicted modes by conditioning on agent goals\cite{zhao2020tnt, mangalam2020not, mangalam2020goals}, lane center-lines\cite{zhang2020map, chang2019argoverse, luo2020probabilistic} or anchor trajectories\cite{chai2020multipath, phan2020covernet}.

Another set of approaches \cite{ziebart2009planning, kitani2012activity, wulfmeier2016watch, zhang2018integrating} pioneered by Ziebart \textit{et al.} \cite{ziebart2009planning}, model agents as Markov decision processes (MDPs) exploring a 2-D grid defined over the scene. A reward map for the MDP is learned via maximum-entropy inverse reinforcement learning (MaxEnt IRL) \cite{ziebart2008maximum}. MDPs are naturally suited to model the agent's sequential decision making. Additionally, since the reward is learned from local scene cues at each grid location, it can be transferred to unknown scenes with a different configuration of scene elements. However, MaxEnt IRL approaches suffer from two limitations: First,
they require a pre-defined absorbing goal state, limiting them to applications where goals of agents are known beforehand. As opposed to this, we need to infer goals of agents. Second, they only provide future locations of the agent in the grid, without mapping them to specific times \footnote{We refer to agent locations without assigned times as \textit{paths}, and agent
locations with assigned times as \textit{trajectories}}. This does not take into account the agent's dynamics.

\textbf{In this work}, we seek to leverage the transferability of grid based MaxEnt IRL approaches, while allowing for sampling of continuous valued trajectories similar to conditional generative models. We present \textit{P2T} (Plans-to-Trajectories), a planning based approach to generate long-term trajectory forecasts in unknown environments. In particular, our approach relies on two key ideas.  
\begin{enumerate}
\item \textbf{Joint inference of goals and paths by learning rewards:} We reformulate the maximum entropy inverse reinforcement learning framework to learn transient path state rewards and terminal goal state rewards. Our reformulation allows for joint inference of goals, and paths to goals. This alleviates the need for a pre-defined absorbing goal state in the original formulation \cite{ziebart2008maximum}. 
\item \textbf{Trajectories conditioned on plans:} We refer to state sequences sampled from the MaxEnt policy as \textit{plans}. We propose an attention based trajectory generator that outputs continuous valued trajectories conditioned on sampled plans, rather than a latent variable.  Compared to conditional generative models, our model outputs trajectories that better conform to the underlying scene over longer prediction horizons. Additionally, the state sequences of the MaxEnt policy allow for better interpretability compared to the latent space of a conditional generative model
\end{enumerate}

Figure \ref{fig_1} shows forecasts generated by P2T for two example scenarios from the NuScenes \cite{caesar2019nuscenes} (top row) and Standord drone \cite{robicquet2016learning} (bottom row) datasets. Our model infers potential goals of the agent and paths to these goals over a coarse 2-D grid defined over the scene. We visualize these goals and paths using \textit{state visitation frequencies} of our MaxEnt policy. We note that the goals and paths conform to the underlying scene, and past motion of the agent, and that the state distribution of the MaxEnt policy is multimodal. We observe that the continuous valued trajectories generated by our model conform to the grid-based plans sampled from the policy. Finally, we cluster the sampled trajectories from the model to give \textit{K} trajectories to be used by a downstream planner.

We evaluate our model on two publicly available trajectory datasets: the Stanford drone dataset \cite{robicquet2016learning} (SDD) consisting of pedestrians, bicyclists, skateboarders and slow moving vehicles at various locations on a university campus, and the NuScenes dataset \cite{caesar2020nuscenes} consisting of vehicles navigating complex urban traffic. We report results in terms of minimum over \textit{K} average displacement error (MinADE$_{K}$), final displacement error (MinFDE$_{K}$) and miss rate (MR$_{K}$) metrics reported in prior work \cite{gupta2018social, lee2017desire, sadeghian2018sophie, rhinehart2019precog, zhao2019multi, chang2019argoverse}, as well as sample quality metrics such as off-road rate \cite{niedoba2019improving} and off-yaw rate \cite{greer2020trajectory}. Our model achieves state of the art results on several metrics, while being competitive on others. In particular, it significantly outperforms existing approaches in terms of sample quality metrics, forecasting trajectories that are both diverse as well as precise. We make our code publicly available at \href{https://github.com/nachiket92/P2T}{https://github.com/nachiket92/P2T}.


\section{Preliminaries}

In this section, we briefly review maximum entropy inverse reinforcement learning (MaxEnt IRL) for path forecasting, conditioned on pre-defined goal states \cite{ziebart2009planning, kitani2012activity, wulfmeier2016watch}.

\vspace{0.1in}
\noindent\textbf{MDP formulation:} We consider a Markov decision process $\mathcal{M} = \{\mathcal{S}, \mathcal{A}, \mathcal{T},r\}$, for a finite horizon setting with $N$ steps. $\mathcal{S}$ is the state space consisting of cells in a 2-D grid defined over the scene. $\mathcal{A}$ is the action space consisting of 4 discrete actions, \textit{\{up, down, left, right\}}, to move to adjacent cells.  We assume deterministic dynamics, where $\mathcal{T}:\mathcal{S}\times\mathcal{A}\to\mathcal{S}$ is the state transition function. Finally, $r: \mathcal{S} \to \mathbb{R}_{0}^{-}$ is the reward function mapping each state to a real value less than or equal to 0. We assume that the initial state $s_{init}$ and the goal state $s_{goal}$ of the MDP are known.

\vspace{0.1in}
\noindent\textbf{MaxEnt IRL objective:} Under the maximum entropy distribution, the probability of observing a state action sequence $\tau=\{(s_1, a_1), (s_2, a_2), \ldots (s_N, a_N)\}$ is proportional to the exponential of its reward.

$$
    P(\tau) = \frac{1}{Z}\mbox{exp}\left(r(\tau)\right) = \frac{1}{Z}\mbox{exp}\left(\sum_{i=1}^{N} r(s_i)\right), \eqno{(1)}
$$
where $Z$ the normalizing constant. MaxEnt IRL involves learning a reward function $r_{\theta}(s)$ parametrized by a set of parameters $\theta$, operating on a set of features extracted for each state $s$. The objective is to learn a reward function that maximizes the log likelihood of observing a training set of demonstrations $\mathrm{T} =  \{\tau_1, \tau_2, \ldots \tau_K \}$ 

$$
 \argmax_{\theta} \mathcal{L_{\theta}} = \argmax_{\theta} \sum_{\tau \in \mathrm{T}} \mbox{log} \left( \frac{1}{Z_{\theta}} \mbox{exp}(r_{\theta}(\tau)) \right). \eqno{(2)}
$$
This can be solved using stochastic gradient descent, with the gradient of the log likelihood $\mathcal{L_{\theta}}$ simplifying to
$$
\frac{d\mathcal{L_{\theta}}}{d\theta} = \sum_{\tau \in \mathrm{T}} (D_{\tau} - D_{\theta}) \frac{d r_{\theta}}{d\theta}, \eqno{(3)}
$$
where, $D_{\tau}$ are the state visitation frequencies (SVFs) for the training demonstration $\tau$ and $D_{\theta}$ are the expected SVFs for the MaxEnt policy given the current set of reward parameters $\theta$. If a deep neural network is used to model the reward function $r_{\theta}(s)$, $\frac{d r_{\theta}}{d\theta}$ can be obtained using backpropagation as described in \cite{wulfmeier2015maximum}. $D_{\theta}$ is obtained using Algorithm \ref{alg:avi_goal} and Algorithm \ref{alg:svf_goal}.

\vspace{0.1in}
\noindent\textbf{Approximate value iteration:} Algorithm \ref{alg:avi_goal} involves solving for the MaxEnt policy $\pi_{\theta}$, given the current reward function $r_{\theta}$, and the goal state $s_{goal}$. $\pi_{\theta}$ represents the probability of taking action $a$, given state $s$. The policy can be stationary, \textit{ie.}, independent of the time step $\pi_{\theta}(a|s)$, or non-stationary $\pi_{\theta}^{(n)}(a|s)$. We use a non-stationary policy as used in \cite{ziebart2010modeling, levine2018reinforcement}. Algorithm \ref{alg:avi_goal} involves iterative updates of the state and action log partition functions $V(s)$ and $Q(s,a)$. These can be interpreted as soft estimates of the expected future reward given state $s$ and the expected future reward given state-action pair $(s,a)$ respectively. $V(s)$ is initialized to 0 for $s_{goal}$ and  $-\infty$ for all other states. $V(s)$ and $Q(s,a)$ are then iteratively updated over $N$ steps, while holding $V(s_{goal})$ fixed at 0. For each step, $\pi_{\theta}$ is given by $$\pi_{\theta}^{(n)}(a|s) = \mbox{exp}\left(Q^{(n)}(s,a)-V^{(n)}(s)\right). \eqno{(4)}$$ Holding $V(s_{goal})$ fixed to 0, while initializing all other $V(s)$ values to $-\infty$ ensures that the MDP ends at $s_{goal}$.

\vspace{0.1in}
\noindent \textbf{Policy propagation:} Algorithm \ref{alg:svf_goal} involves calculating the SVFs. It involves repeatedly applying $\pi_{\theta}$ for $N$ steps, starting with the initial state distribution, to give SVF at each step. The SVF corresponding to the goal state is set to 0 at each step, since the goal state absorbs any probability mass that reaches it. The expected SVF $D_{\theta}$ is obtained by summing the SVFs over the $N$ steps.

{\centering
\begin{minipage}{\linewidth}
\begin{algorithm}[H]
  \caption{Approx. value iteration (goal conditioned)}\label{alg:avi_goal}
  \hspace*{\algorithmicindent}\textbf{Inputs:} $r_{\theta}$, $s_{goal}$
  \begin{algorithmic}[1]
  \State $V^{(N)}(s) \gets -\infty, \; \forall s \in \mathcal{S}$               
  \For{$n = N,...,2,1$}
    \State $V^{(n)}(s_{goal}) \gets 0 $
    \State $Q^{(n)}(s,a) = r_{\theta}(s) + V^{(n)}(s'),  \;\;\;\;\;\;\;\;\;\;\;\;\;\; s'= \mathcal{T}(s,a)$
    \State $V^{(n-1)}(s) = \mbox{logsumexp}_{a}$ $Q^{(n)}(s,a)$ 
    \State $\pi_{\theta}^{(n)}(a|s) =  \mbox{exp}\left(Q^{(n)}(s,a)-V^{(n)}(s)\right)$
  \EndFor

  \end{algorithmic}
\end{algorithm}
\end{minipage}
\par
}

{\centering
\begin{minipage}{\linewidth}
\begin{algorithm}[H]
  \caption{Policy propagation (goal conditioned)}\label{alg:svf_goal}
  \hspace*{\algorithmicindent}\textbf{Inputs:} $\pi_{\theta}$, $s_{init}$, $s_{goal}$
  \begin{algorithmic}[1]
  \State $D^{(1)}(s) \gets 0 , \; \forall s \in \mathcal{S} $ 
  \State $D^{(1)}(s_{init}) \gets 1$               
  \For{$n = 1,2...,N$}
    \State $D^{(n)}(s_{goal}) \gets 0 $
    \State $D^{(n+1)}(s) = \sum_{s', a} \pi_{\theta}^{(n)}(a|s')D^{(n)}(s'), \;\;\; \mathcal{T}(s',a)=s$
  \EndFor
  \State $D(s) = \sum_n D^{(n)}(s)$
    
  \end{algorithmic}
\end{algorithm}
\end{minipage}
\par
}

\vspace{0.1in}
\noindent \textbf{Path forecasting conditioned on goals:} The MaxEnt policy $\pi_{\theta}^{*}$, for the converged reward model $r_{\theta}$, can be sampled from, to give path forecasts on the 2-D grid from the $s_{init}$ to $s_{goal}$. Since $\pi_{\theta}^{*}$ is stochastic, the policy can explore multiple paths within the scene to the goal state. However, for most cases of pedestrian or vehicle trajectory forecasting, $s_{goal}$ is unknown, and needs to be inferred. Additionally, sampling $\pi_{\theta}^{*}$ only provides future paths, without mapping them to specific times. A step for the MDP need not correspond to a fixed time interval. Different agents can have different speeds. Agents can also accelerate or decelerate over the course of the 10s prediction horizon.

\begin{figure*}[t]
\centering
\includegraphics[width=\linewidth]{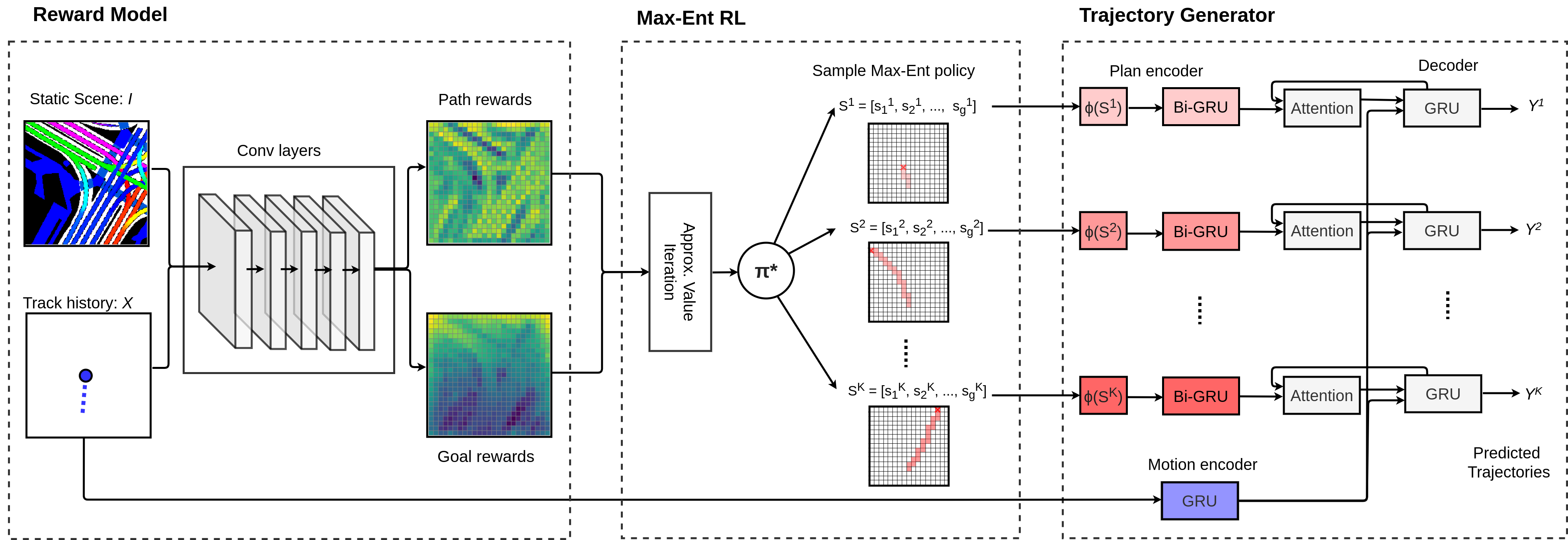}
\caption{\textbf{Overview}: P2T consists of three modules: (1) a fully convolutional reward model, that outputs transient path state rewards and terminal goal state rewards on a coarse 2-D grid, (2) a MaxEnt RL policy for the learned path and state rewards, that can be sampled to generate multimodal plans on the 2-D grid, and (3) an attention based trajectory generator, that outputs continuous valued trajectories conditioned on the sampled plans.}
\label{fig_2}
\end{figure*}

\section{Proposed Approach}
\label{sec:model}
We leverage the transferability of grid based MaxEnt IRL, while not requiring knowledge of $s_{goal}$, and generate continuous valued trajectories, mapped to specific times in the future. Figure \ref{fig_2} provides an overview of P2T, our proposed approach. P2T consists of three components. 

The first component is a reward model, comprised by convolutional and pooling layers. At each cell on a coarse 2-D grid, the reward model maps local scene context and motion features capturing the agent's track history, to a transient path state reward and a terminal goal state reward. We describe the reward model in greater detail in section \ref{sec:reward_model}. 

The next component is a MaxEnt policy independent of pre-defined goal states. We reformulate MaxEnt IRL to allow for inference of goal and path states, given the path and goal rewards learned by the reward model (see section \ref{sec:maxentirl_changes}). We obtain a single policy that can be sampled to generate paths to different plausible goals on the 2-D grid. We refer to each state sequence sampled from the policy as a \textit{plan}.

The final component of P2T is an attention based trajectory generator, that outputs continuous valued trajectories conditioned on the sampled plans. The trajectory generator encodes the track history of the agent using a gated recurrent unit (GRU), and the sampled plans using a bidirectional GRU (BiGRU). Finally, a GRU decoder equipped with soft-attention \cite{bahdanau2014neural}, attends to the plan encoding to output trajectories over the prediction horizon. Section \ref{sec:traj_gen} describes the trajectory generator in greater detail.

\subsection{Inferring goals and paths by learning rewards}\label{sec:maxentirl_changes}
We wish to relax the requirement of prior knowledge of $s_{goal}$ in MaxEnt IRL. Certain locations in a scene are likelier to be goals of agents. For pedestrians, these can be points where paths and sidewalks exit the scene, entrances to buildings, or parked cars. For vehicles, these can be points where lanes exit the scene, stop signs or parking lots. Goals are also likelier to be along the direction of the agent's motion. Rather than always terminating at a predefined goal, we would like our policy to induce a distribution over possible goal states. This would allow us to sample paths from the policy terminating at different goals in the scene. We propose to do this by learning path and goal state rewards, conditioned on the scene and past motion of the agent, and learning a policy unconstrained by $s_{goal}$. We reformulate the MDP and modify the approximate value iteration algorithm.

\vspace{0.15in}

\noindent\textbf{MDP formulation:}
\begin{itemize}
\vspace{0.05in}

    \item \textbf{State space:} Potentially any cell location on the 2-D grid could be the goal of the agent, or a point on their future path. We define the state space $\mathcal{S} = \left\{\mathcal{S}_p, \mathcal{S}_g  \right\}$. $\mathcal{S}_p$ is the set of path states and $\mathcal{S}_g$ is the set of goal states. Each cell location on the 2-D grid has an associated path state belonging to $\mathcal{S}_p$ and a goal state belonging to $\mathcal{S}_g$. The policy terminates on reaching any goal state.   
    
    \vspace{0.05in}
    \item \textbf{Action space:} $\mathcal{A}$ = \{\textit{up, down, left, right, end}\}. The \textit{up}, \textit{down}, \textit{left} and \textit{right} actions allow transitions from path states to adjacent path states. Additionally, we define an \textit{end} action that transitions the MDP from a path state to the goal state at the same cell location.  
    \vspace{0.05in}
    \item \textbf{Transition function:} $\mathcal{T}:\mathcal{S}_p \times \mathcal{A} \to \mathcal{S}$ maps path state and action pairs to other path states and goal states. Since goal states are terminal, the MDP has no transitions out of a goal state.  
    
    \vspace{0.05in}
    \item \textbf{Rewards:} We learn two functions, $r_{p_{\theta}}$ corresponding to path rewards, and $r_{g_{\theta}}$ corresponding to goal rewards. 
\end{itemize}

\noindent\textbf{Approximate value iteration with inferred goals:}
Algorithm \ref{alg:avi} depicts our modified approximate value iteration, unconstrained on $s_{goal}$. Unlike algorithm \ref{alg:avi_goal}, we do not hold the $V(s_{goal})$ fixed at 0 to enforce goal directed behavior. Instead, we use $r_{g_{\theta}}$ to learn a policy that induces a multimodal distribution over potential goal states. The inputs to algorithm \ref{alg:avi} are the learned rewards $r_{g_{\theta}}$ and $r_{p_{\theta}}$. We initialize $V(s)$ to $-\infty$ for all path states $\mathcal{S}_p$. This is because we want the MDP to end up at some goal state within the $N$ step finite horizon. Since the goal states are terminal, the MDP receives the goal rewards only once. We thus hold $V(s)$ fixed to $r_{g_{\theta}}(s)$ for all goal states $\mathcal{S}_g$. We then iteratively update the state-action log partition function $Q^{(n)}(s,a)$ and the state log partition function $V^{(n)}(s)$ for the path states $\mathcal{S}_p$ over $N$ steps. At the end of each step, the MaxEnt policy is obtained by taking the ratio of the exponent of $Q^{(n)}(s,a)$ and $V^{(n)}(s)$.

\vspace{0.1in}
\noindent\textbf{Policy propagation with inferred goals:} Algorithm \ref{alg:svf} depicts policy propagation independent of $s_{goal}$. This is almost identical to algorithm \ref{alg:svf_goal}. The only difference is, we do not set the goal state SVFs to 0, as in line 4 of algorithm \ref{alg:svf_goal}. This is because we use the goal SVFs to train the reward model for $r_{g_{\theta}}$, using equation (3). We use a frame of reference centered at the agent's location at the time of prediction. Thus, $s_{init}$ is always the path state at the center of the grid.

{\centering
\begin{minipage}{\linewidth}
\begin{algorithm}[H]
  \caption{Approx. value iteration (inferred goals)}\label{alg:avi}
  \hspace*{\algorithmicindent}\textbf{Inputs:} $r_{g_{\theta}}$, $r_{p_{\theta}}$ 
  \begin{algorithmic}[1]
  \State $V^{(N)}(s) \gets -\infty, \; \forall s \in \mathcal{S}_p$               
  \For{$n = N,...,2,1$}
    \State $V^{(n)}(s) \gets r_{g_{\theta}}(s), \; \forall s \in \mathcal{S}_g $
    \State $Q^{(n)}(s,a) = r_{p_{\theta}}(s) + V^{(n)}(s'),  \; \forall s \in \mathcal{S}_p, \; s'= \mathcal{T}(s,a)$
    \State $V^{(n-1)}(s) = \mbox{logsumexp}_{a}$ $Q^{(n)}(s,a),  \; \forall s \in \mathcal{S}_p $ 
    \State $\pi_{\theta}^{(n)}(a|s) =  \mbox{exp}\left(Q^{(n)}(s,a)-V^{(n)}(s)\right)$
  \EndFor

  \end{algorithmic}
\end{algorithm}
\end{minipage}
\par
}

{\centering
\begin{minipage}{\linewidth}
\begin{algorithm}[H]
  \caption{Policy propagation (inferred goals)}\label{alg:svf}
  \hspace*{\algorithmicindent}\textbf{Inputs:} $\pi_{\theta}$, $s_{init}$
  \begin{algorithmic}[1]
  \State $D^{(1)}(s) \gets 0 , \; \forall s \in \mathcal{S} $ 
  \State $D^{(1)}(s_{init}) \gets 1$               
  \For{$n = 1,2...,N$}
    \State $D^{(n+1)}(s) = \sum_{s', a} \pi_{\theta}^{(n)}(a|s')D^{(n)}(s'), \;\;\; \mathcal{T}(s',a)=s$
  \EndFor
  \State $D(s) = \sum_n D^{(n)}(s)$
    
  \end{algorithmic}
\end{algorithm}
\end{minipage}
\par
}

\subsection{Reward model}\label{sec:reward_model}
\begin{figure}[t]
\centering
\includegraphics[width=\columnwidth]{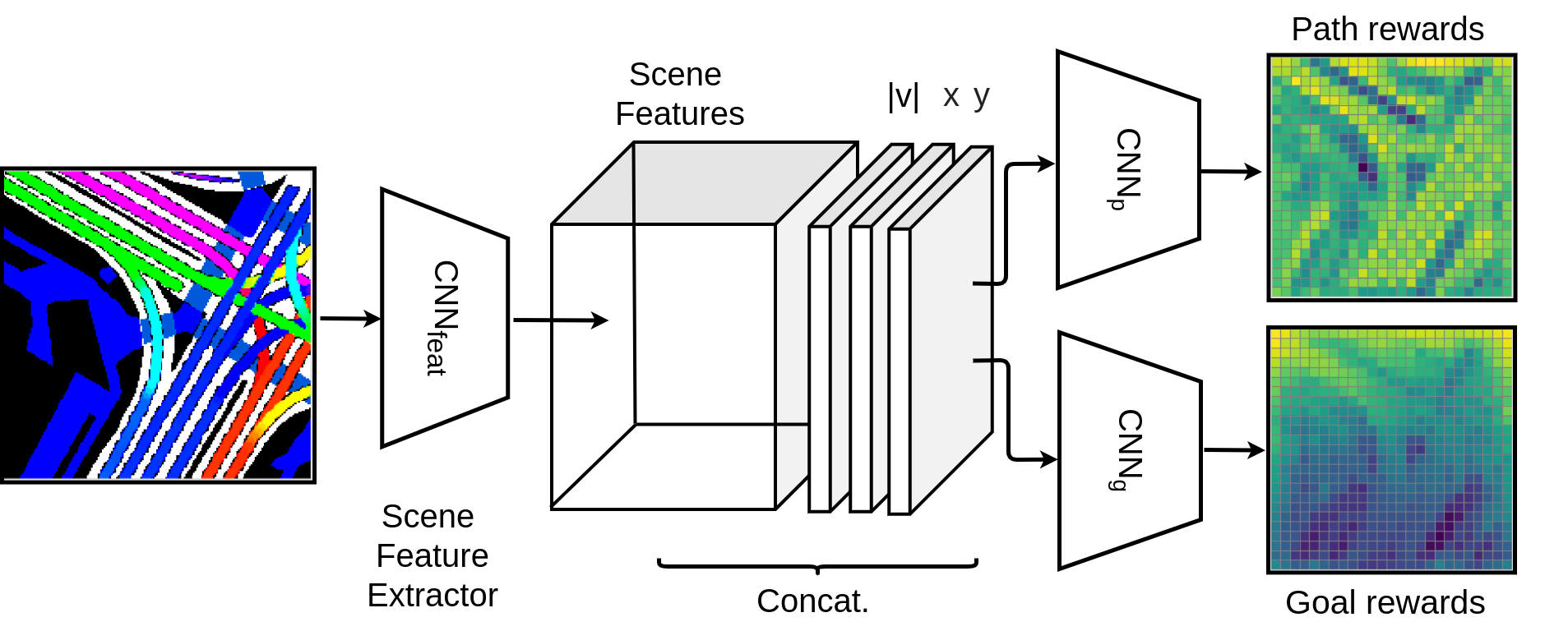}
\caption{\textbf{Reward model}: $\mbox{CNN}_{feat}$ extracts features from the static scene. We concatenate these with feature maps capturing the agent's motion. $\mbox{CNN}_p$ and $\mbox{CNN}_g$ learn path and goal rewards from the features.}
\label{fig:reward_model}
\end{figure}

We define a reward model consisting purely of convolutional and pooling layers. This allows us to learn a mapping from local patches of the scene to path and goal rewards. The equivariance of the convolutional layers allows the reward model to be transferred to novel scenes with a different configuration of scene elements. Figure \ref{fig:reward_model} shows our reward model. It consists of three sets of convolutional layers. 

$\mbox{CNN}_{feat}$ serves as a scene feature extractor, operating on the birds eye view representation $I$ of the static scene around the agent: 
$$\phi_{I} = \mbox{CNN}_{feat}\left(I\right). \eqno{(5)}$$
The spatial dimensions of the scene features $\phi_I$ equal the size of the 2-D grid corresponding to our state space $\mathcal{S}$. In addition to scene features, we want our goal and path rewards to depend on the past motion of the agent. Thus, similar to Zhang \textit{et al.} \cite{zhang2018integrating}, we concatenate the scene features with feature maps encoding the agent's motion, and the locations of the grid cells:
$$\phi_{M} = \left[|v|, x, y \right]. \eqno{(6)}$$
Here, $|v|$ is the speed of the agent. This value is replicated over the entire feature map. $x$ and $y$ are the locations of each grid cell in the agent-centric frame of reference, with the origin at the agent's current location and the x-axis aligned along the agent's current direction of motion. 

$\mbox{CNN}_p$ and $\mbox{CNN}_g$ map the scene and motion features to path and goal rewards respectively:
$$r_{p_{\theta}} = \mbox{CNN}_p\left(\phi_{I}, \phi_{M}\right). \eqno{(7)}$$
$$r_{g_{\theta}} = \mbox{CNN}_g\left(\phi_{I}, \phi_{M}\right). \eqno{(8)}$$

\noindent\textbf{Implementation details:}
We represent the scene as a $200\times200$ bird's eye view image around the agent. $\mbox{CNN}_{feat}$ consists of the first two ImageNet pretrained blocks of ResNet34 \cite{he2016deep}. This downsamples the spatial dimension of the feature maps to $50\times50$. This is followed by a $2\times2$ convolutional layer with depth 32 and stride 2, to aggregate context at each cell location. This gives 32 scene feature maps over a $25\times25$ grid. $\mbox{CNN}_p$ and $\mbox{CNN}_g$ have identical architectures consisting of two $1\times1$ convolutional layers. The first layer has depth 32, and the second layer has depth 1 to give a single path or goal reward value at each cell. We apply the log-sigmoid activation at the outputs of $\mbox{CNN}_p$ and $\mbox{CNN}_g$ to restrict reward values between $-\infty$ and 0. The reward model is trained to maximize the log-likelihood $\mathcal{L}_{\theta}$ of agent paths in the train set shown in equation (2), with gradients given by equation (3). The state visitation frequencies $D_{\theta}$ for both path and goal states are obtained using algorithms \ref{alg:avi} and \ref{alg:svf}. We use Adam \cite{kingma2014adam} with learning rate 0.0001 to train the reward model.

\subsection{Trajectories conditioned on plans}\label{sec:traj_gen}

Consider the optimal MaxEnt policy $\pi_{\theta}^{*}$ obtained using algorithm \ref{alg:avi} for the converged reward model. Consider state sequences or \textit{plans} sampled from $\pi_{\theta}^{*}$, with the $i^{th}$ plan given by
$$s^{(i)} = \left[s_{1}^{(i)}, s_{2}^{(i)}, \dots, s_{N}^{(i)} \right]. \eqno{(9)}$$
We expect the sampled plans to end at a diverse set of goal states, and explore various paths to these goals. Additionally, each plan $S^{(i)}$ can be expected to conform to the underlying scene and model the agent's sequential decision making. However, the plans by themselves do not capture the dynamics of the agent's motion. A fast moving agent can make more progress along a plan compared to a slow moving agent over a fixed prediction horizon $T_f$. The dynamics of the agent's motion can be estimated using a snippet of their most recent track history, over time $T_h$,
$$x = \left[x_{-T_h}, \dots, x_{1}, x_{0}\right], \eqno{(10)}$$
where the $x_t$'s correspond to past location, velocity, acceleration and yaw-rate of the agent, with the subscript $t$ representing time and $t=0$ the prediction instant.

We thus seek a model that, for each sampled plan $s^{(i)}$, and track history $x$, generates a continuous valued trajectory $y^{(i)}$ over a prediction horizon $T_f$,
$$y^{(i)} = \left[y_{1}^{(i)}, y_{2}^{(i)}, \dots, y_{T_f}^{(i)}\right], \eqno{(11)}$$
where $y_t$ is the future location of the agent at time $t$. We propose a trajectory generator modeled as a recurrent neural network encoder-decoder, equipped with soft attention \cite{bahdanau2014neural}. The trajectory generator has the following components.

\begin{figure}[t]
\centering
\includegraphics[width=\columnwidth]{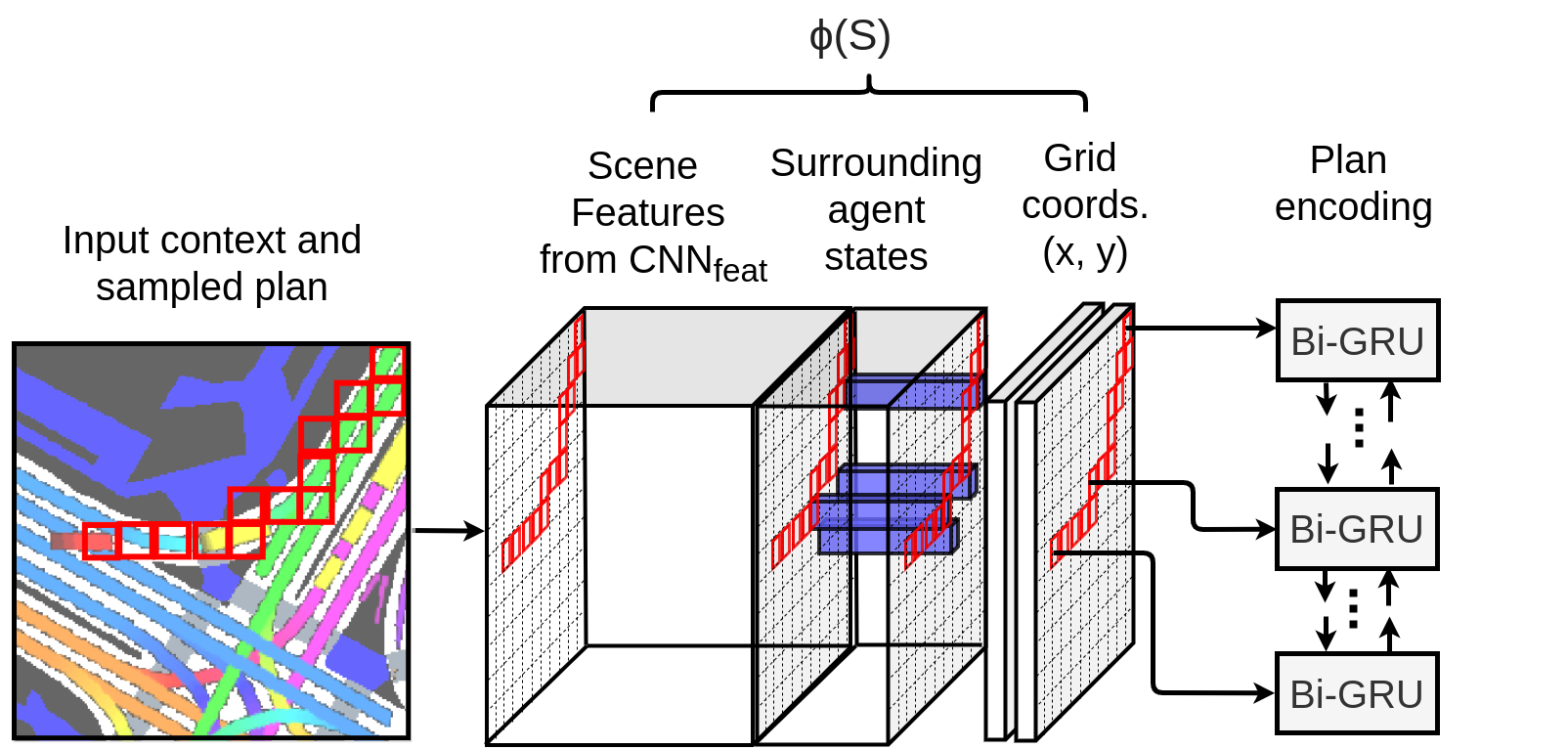}
\caption{\textbf{Plan encoder}: For each state in a sampled plan, we encode the scene features, surrounding agent states and the location co-ordinates of the grid cell and term it $\phi_{S}(s)$. This is then fed into bidirectional GRU to encode the the entire sampled plan. Our GRU decoder generates output trajectories by attending to the plan encoding.}
\label{fig:plan_encoder}
\end{figure}

\begin{itemize}

\vspace{0.05in}
\item \noindent\textbf{Motion encoder:} We encode the track history $x$ using a GRU encoder, where the state of the GRU at time $t$ is given by
$$h_{m_t} = \mbox{GRU}_m\left(h_{m_{t-1}}, e_x\left(x_t\right) \right). \eqno{(12)}$$ 
Here $e_x(\cdot)$ is a fully connected embedding layer. The GRU state at the prediction instant, $h_{m_0}$, can be expected to encode the motion of the agent.

\vspace{0.05in}
\item \noindent\textbf{Plan encoder:}
The plan encoder (Fig. \ref{fig:plan_encoder}) encodes local scene features, nearby agent states and location co-ordinates along sampled state sequences. The plan encoding serves as an inductive bias for the decoder to output trajectories that conform to the paths and goals inferred by the policy. For scene features, we use the outputs $\phi_I$ of CNN$_{feat}$ from the reward model (equation 5). For surrounding agent states, we populate their grid locations with the agents' velocity, acceleration and yaw-rate. For each state $s_n^{(i)}$ in a sampled plan $s^{(i)}$, we embed the scene features, agent states and location co-ordinates at the grid cell corresponding to $s_n^{(i)}$, using fully connected layers and concatenate the outputs to give the state encoding $\phi_s\left(s_{n}^{(i)}\right)$. We use a bidirectional GRU (BiGRU) encoder to aggregate the state encodings over the entire plan. The state of the BiGRU at step $n$ is given by   
$$h_{s_n}^{(i)} = \mbox{BiGRU}_s\left(h_{s_{n-1}}^{(i)}, h_{s_{n+1}}^{(i)}, \phi_s\left(s_{n}^{(i)}\right) \right). \eqno{(13)}$$

\vspace{0.05in}
\item \noindent\textbf{Attention based decoder:} We use a GRU decoder equipped with a soft attention module to generate the output trajectories $y^{(i)}$. Our core idea is to allow the decoder to attend to specific states of the sampled plan $s^{(i)}$ as it generates trajectories along the plan. Thus, the decoder can attend to just the first few states of sampled plans, as it generates the future trajectories for a slow moving agent. On the other hand, it can attend to later states while generating a fast moving agent's trajectories.   

We initialize the state of the decoder using the final state of the motion encoder,
$$ h_{dec_1} = h_{m_0}. \eqno{(14)}$$
This provides the decoder a representation of the agent's motion. The decoder state is then updated over the prediction horizon, with the outputs at each time-step giving the predicted locations.

$$h_{dec_t}^{(i)} = \mbox{GRU}_{dec}\left(h_{dec_{t-1}}^{(i)}, \mbox{Att}\left(h_{dec_{t-1}}^{(i)}, h_{s_{1:N}}^{(i)}\right)    \right),  \eqno{(15)}$$

$$y_t^{(i)} = o_y(h_{dec_t}^{(i)}), \eqno{(16)} $$
where $\mbox{Att}(\cdot)$ is the attention module and $o_y(\cdot)$ is a fully connected layer operating on the decoder states. 

\end{itemize}

\vspace{0.1in}
\noindent \textbf{Sampling and clustering:} The trajectory generator outputs a trajectory conditioned on each sampled plan. This allows us to indefinitely sample trajectories from our model. Since the MaxEnt policy induces a multimodal distribution over path and goal states, the sampled trajectories also represent a multimodal predictive distribution. However, sampling by itself can be inefficient, with several sampled state sequences and trajectories being identical or very similar. In order to provide downstream path planners with a succinct representation of the trajectory distribution, we cluster the sampled trajectories using the K-means algorithm to output a set of $K$ predicted trajectories. The number of clusters $K$ can be varied as required by the downstream path planner, without having to re-train the model.

\vspace{0.1in}
\noindent\textbf{Implementation details:} As per the standard benchmarks for both datasets, we use track history of 3.2 seconds and a prediction horizon of 4.8 seconds SDD \cite{robicquet2016learning}, while a track history of 2 seconds and a prediction horizon of 6 seconds for NuScenes \cite{caesar2020nuscenes}. We assume an agent centric frame of reference with the x-axis aligned along the agent's direction of motion at $t=0$. We use a 32 sized state vector for each of the GRUs. The motion encoder uses an embedding layer of size 16, while the plan encoder uses embedding layers of size 16, 32 and 16 for grid locations, scene features and surrounding agent states respectively. Our attention module is a multi-layer perceptron (MLP) with one hidden layer of size 32. 

To train the model, we sample 200 plans and corresponding trajectories from the trajectory generator and cluster them to give $K$ output trajectories $\left[y^{(1)}, y^{(2)},\dots,y^{(K)}\right]$. We use $K=10$ for NuScenes and $K=20$ for SDD. We minimize the minimum over $K$ average displacement error (MinADE$_K$) over the training set.
$$\mbox{MinADE}_K = \min_{i \in \left\{1,\dots,K\right\}} \frac{1}{T_f} \sum_{t=1}^{T_f}\left\|y^{GT}_t - y^{(i)}_t\right\|_{2}, \eqno{(17)}$$
where $y^{GT}$ is the ground truth future trajectory of the agent. The MinADE$_K$ loss has been used in prior work for training models for multimodal trajectory forecasting \cite{gupta2018social, cui2019multimodal, sadeghian2018sophie}. For a model generating multiple trajectories, it avoids penalizing plausible future trajectories that do not correspond to the ground truth. 
To speed up convergence, we pre-train the model to minimize the average displacement error between $y^{GT}$ and the trajectory predicted by the model conditioned on the ground truth plan of the agent $y^{S_{GT}}$. We use Adam, with learning rate 0.0001 for training the trajectory generator

\section{Experimental Analysis}

\subsection{Datasets}

\noindent \textbf{Stanford drone dataset:}  The Stanford drone dataset (SDD) \cite{robicquet2016learning} consists of trajectories of pedestrians, bicyclists, skateboarders and vehicles captured using drones at 60 different scenes on the Stanford university campus. The dataset provides bird's eye view images of the scenes, and locations of tracked agents in the scene's pixel co-ordinates. The dataset contains a diverse set of scene elements like roads, sidewalks, walkways, buildings, parking lots, terrain and foliage. The roads and walkways have different configurations, including roundabouts and four-way intersections. We use the dataset split defined in the TrajNet benchmark \cite{sadeghian2018trajnet} and used in prior work \cite{sadeghian2018sophie, zhao2019multi, bhattacharyya2019conditional}, for defining our train, validation and test sets. The dataset is split based on scenes. Thus, the train, validation and test sets all have different scenes from the 60 total scenes. This allows us to evaluate our model on unknown scenes where it hasn't seen prior trajectory data. Note that we consider \textit{all trajectories} in the train, validation and test scenes of SDD as per \cite{sadeghian2018sophie, zhao2019multi, bhattacharyya2019conditional, bhattacharyya2020haar}. Subsequent work \cite{mangalam2020goals, mangalam2020not} has reported results on a subset of trajectories primarily consisting of pedestrians. We report results on this split separately.     

\vspace{0.1in}
\noindent \textbf{NuScenes:} 
The NuScenes dataset \cite{caesar2020nuscenes} comprises 1000 scenes, each of which is a 20 second record, capturing complex urban traffic over a wide variety of road layouts and lane structures. The dataset was captured using vehicle mounted camera and lidar sensors while driving through Boston and Singapore, and contains hand annotated vehicle detection boxes and tracks at a 2 Hz. In particular, we train and evaluate our model using the official benchmark split for the NuScenes prediction challenge consisting of vehicle trajectories. In addition to trajectories, NuScenes provides high definition bird's eye view maps of the scene, including drivable area masks, cross-walks, side-walks and lane center-lines along with their connectivity and directions. We use a 50m $\times$ 50m crop of the HD map around the vehicle of interest as the scene representation for our model. It extends 40m along the agent's direction of motion, 10m behind and $\pm$ 25m laterally.

\subsection{Metrics}

\begin{figure}[t]
\centering
\includegraphics[width=\columnwidth]{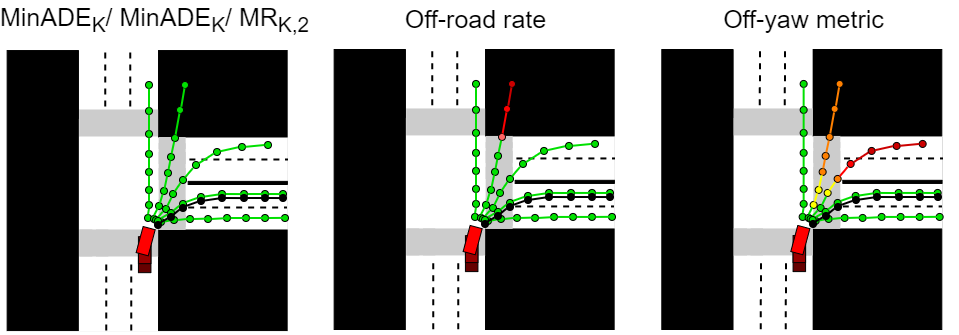}
\caption{\textbf{Sample quality metrics.} MinADE$_K$, MinFDE$_K$ and miss rate fail to penalize a diverse set of trajectories that don't conform to the scene (left). The off-road rate (middle) and off-yaw (right) metrics address this by penalizing predicted points that fall off the drivable area or onto oncoming traffic. Warm colors indicate higher errors.} 
\label{fig:metrics}
\end{figure}

\begin{table*}[]
\caption{Results on SDD test set for split used in \cite{sadeghian2018sophie}}
\centering
\begin{tabularx}{0.7\textwidth}{lYYYYY}
\toprule
Model & MinADE$_5$ & MinADE$_{20}$ & MinFDE$_5$ & MinFDE$_{20}$ &  \begin{tabular}[c]{@{}c@{}}Off-road\\ rate\end{tabular}
\\ \midrule

S-GAN \cite{gupta2018social}     & -                     & 27.25                   & -                     & 41.44                   & -                                                       \\
Desire \cite{lee2017desire}    & 19.25                  & -                      & 34.05                  & -                      & -                                                       \\
MATF \cite{zhao2019multi}  & -                     & 22.59                   & -                     & 33.53                   & -                                                       \\
SoPhie \cite{sadeghian2018sophie}    & -                     & 16.27                   & -                     & 29.38                   & -                                                       \\
CF-VAE \cite{bhattacharyya2019conditional}    & -                     & 12.60                   & -                     & 22.30                   & -                                                       \\
HBA-flow \cite{bhattacharyya2020haar}  & -                     & \textbf{10.80}                   & -                     & 19.80                   & -                                                       \\ 
P2T (ours) & \textbf{15.90}                 & 10.97                    & \textbf{30.48}                  & \textbf{18.40}                   & 0.06                                                    \\ \bottomrule
\end{tabularx}
\label{tab:sdd1}
\end{table*}

\begin{table*}[]
\caption{Results on SDD test set for split used in \cite{mangalam2020not}}
\centering
\begin{tabularx}{0.7\textwidth}{lYYYYY}
\toprule
Model & MinADE$_5$ & MinADE$_{20}$ & MinFDE$_5$ & MinFDE$_{20}$ &  \begin{tabular}[c]{@{}c@{}}Off-road\\ rate\end{tabular}
\\ \midrule
PECNet \cite{mangalam2020not}      & 12.79                     & 9.96                    & 29.58                     & 15.88                   & -                                                       \\
Y-Net \cite{mangalam2020goals}    & \textbf{11.49}                  & \textbf{7.85}                      & \textbf{20.23}                  & \textbf{11.85}                      & -                                                       \\
P2T (ours) & 12.81                  & 8.76                   & 23.95                  & 14.08                   & 0.06                                                    \\ \bottomrule
\end{tabularx}
\label{tab:sdd2}
\end{table*}

\begin{table*}[]
\caption{Results on NuScenes test set for the prediction benchmark split}
\centering
\begin{tabularx}{0.85\textwidth}{lYYYYYY}
\toprule
Model & MinADE$_5$ & MinADE$_{10}$ & MR$_{5,2}$ & MR$_{10,2}$ &  \begin{tabular}[c]{@{}c@{}}Off-road\\ rate\end{tabular} & \begin{tabular}[c]{@{}c@{}}Off-yaw\\ metric\end{tabular}                                                                        \\ \midrule
Physics oracle\cite{phan2020covernet}            & 3.70       & 3.70         & 0.88       & 0.88      & 0.12                                                                   & -                                                                         \\
CoverNet \cite{phan2020covernet}                 & 2.62        & 1.92         & 0.76      & 0.64      & 0.13                                                                     & -                                                                      \\
MTP \cite{cui2019multimodal}                      & 2.44        & 1.57           & 0.70         & 0.55      & 0.11                                                                     & 0.11                                                                      \\
Trajectron ++ (3rd place) \cite{salzmann2020trajectron++} & 1.88    & 1.51          & 0.70      & 0.57      & 0.25                                                                     & -                                                                      \\
MHA-JAM (2nd place) \cite{messaoud2020multi}       & 1.81    & 1.24          & \textbf{0.59}      & \textbf{0.46}      & 0.07                                                                     & -                                                                      \\
cxx (winning entry) \cite{luo2020probabilistic}      & 1.63    & 1.29          & 0.69      & 0.60      & 0.08                                                                     & -                                                                      \\

Multipath \cite{chai2020multipath}                 & 1.63       & 1.50          & 0.75       & 0.74      & 0.38                                                                     & 0.37                                                                      \\
Noah (current best entry)                 & 1.59       & 1.37          & 0.69       & 0.62      & 0.08                                                                     & 0.37                                                                      \\

P2T (Ours)                       & \textbf{1.45}        & \textbf{1.16}         & 0.64         & \textbf{0.46}      & \textbf{0.03}                                                                     & \textbf{0.04}                                                                      \\ \bottomrule
\end{tabularx}
\label{tab:ns}
\end{table*}

\noindent\textbf{Deviation from ground-truth:} For evaluating a trajectory forecasting model, we need a metric for how much the forecasts deviate from the ground truth future trajectory. However, since our model generates forecasts from a multimodal distribution, we need a metric that does not penalize plausible trajectories generated by the model that don't correspond to the ground truth. Thus, we use the minimum of $K$ average displacement error (MinADE$_K$), final displacement error (MinFDE$_K$) and miss rate within 2 meters (MR$_{K,2}$) as metrics, as utilized in prior work on multimodal trajectory forecasting \cite{gupta2018social, sadeghian2018sophie, lee2017desire, rhinehart2019precog, zhao2019multi, bhattacharyya2019conditional, chang2019argoverse}. MinADE$_K$ (eq. 18) computes the average prediction error in terms of L2 norm between the ground truth future trajectory, and the forecast trajectory closest to it. MinFDE$_K$ is similar to MinADE$_K$, but only considers the  prediction error  for  the  final  predicted  location. Finally, a set of $K$ predictions is considered a missed prediction if none of the $K$ trajectories are within 2 meters of the ground truth over the entire prediction horizon. MR$_{K,2}$ computes the fraction of missed predictions in the test set.

\vspace{0.1in}
\noindent\textbf{Sample quality metrics:} While MinADE$_K$, MinFDE$_K$ and MR$_{K,2}$ avoid penalizing plausible future trajectories that don't conform to the ground truth, they also do not penalize implausible future trajectories as long as one of the $K$ trajectories is close to the ground truth. Thus a model that generates a very diverse set of $K$ trajectories by random guessing can achieve low MinADE$_K$, MinFDE$_K$ and MR$_{K,2}$ values, even if the trajectories do not conform to the underlying scene. Thus, while these metrics serve as good measures of the 'recall' of the model for the multimodal predictive distribution, they serve as poor measures for the model's 'precision'. We refer readers to Rhinehart \textit{et al.} \cite{rhinehart2018r2p2} for a detailed discussion on the diversity-precision trade-off. To evaluate the precision of trajectories generated by our model, we additionally report results on two recently proposed sample quality metrics.
\vspace{0.05in}
\begin{itemize}
    \item \textbf{Off-road rate:} The off-road rate metric proposed by Niedoba \textit{et al.} \cite{niedoba2019improving} computes the fraction of all predicted points that fall outside the road. For the NuScenes dataset, we use the drivable area mask to compute off-road rate. For SDD, we hand label the bird's eye view images in the test set, assigning each pixel as a path or an obstacle. Paths include roads, sidewalks, walkways etc., while obstacles include buildings, terrain, parked cars and road dividers. 
    \vspace{0.05in}
    \item \textbf{Off-yaw metric:} For vehicles moving through city streets, the off-road rate metric fails to penalize predictions that fall onto oncoming traffic or illegal lanes. We thus additionally report the off-yaw metric proposed by Greer \textit{et al.} \cite{greer2020trajectory}, for the NuScenes dataset. The off-yaw metric computes the deviation between the direction of motion of the nearest lane and the yaw of predicted points. Similar to Greer \textit{et al.}, we only penalize deviations above $45 ^{\circ}$ to avoid penalizing lane changes.  
\end{itemize}
\vspace{0.05in}
\noindent Figure \ref{fig:metrics} illustrates how the off-road rate and off-yaw rate can penalize a set of diverse but imprecise forecasts that MinADE$_K$, MinFDE$_K$ and MR$_{K,2}$ metrics fail to penalize.

\begin{table*}[]
\caption{Ablations on SDD}
\centering
\begin{tabularx}{0.9\textwidth}{l|YYYY|YYYYY}
\toprule
Model     & CNN$_{feat}$ & \begin{tabular}[c]{@{}c@{}}Reward\\ layers\end{tabular} & \begin{tabular}[c]{@{}c@{}}Grid-based\\ plans\end{tabular} & \begin{tabular}[c]{@{}c@{}}Trajectory\\ generator\end{tabular} & MinADE$_5$ & MinADE$_{20}$ & MinFDE$_5$ & MinFDE$_{20}$ & \begin{tabular}[c]{@{}c@{}}Offroad\\ rate\end{tabular} \\ \midrule

LVM                    & \ding{51}                     &                                                                          &                                                                                 & \ding{51}                                                                            & 18.28         & 12.17         & 36.71       & 20.98      & 0.11                                                                                                                                          \\
P2T$_{CS}$                  & \ding{51}                     & \ding{51}                                                                     & \ding{51}                                                                            &                                                                                & 21.70         & 16.10          & 38.25       & 25.22   & 0.09                                                                                                                                          \\
P2T$_{BC}$                 & \ding{51}                     &                                                                          & \ding{51}                                                                            & \ding{51}                                                                            & 15.93         & 11.56         & \textbf{30.29}       & 19.51      & \textbf{0.06}                                                                                                                                          \\
P2T$_{IRL}$                & \ding{51}                     & \ding{51}                                                                     & \ding{51}                                                                            & \ding{51}                                                                            & \textbf{15.90}         & \textbf{10.97}         & 30.48       & \textbf{18.40}      & \textbf{0.06}                                                                                                                                        \\ \bottomrule
\end{tabularx}
\label{tab:abl_sdd}
\end{table*}

\begin{table*}[]
\caption{Ablations on NuScenes}
\centering
\begin{tabularx}{\textwidth}{l|YYYY|YYYYYY}
\toprule
Model     & CNN$_{feat}$ & \begin{tabular}[c]{@{}c@{}}Reward\\ layers\end{tabular} & \begin{tabular}[c]{@{}c@{}}Grid-based\\ plans\end{tabular} & \begin{tabular}[c]{@{}c@{}}Trajectory\\ generator\end{tabular} & MinADE$_5$ & MinADE$_{10}$ & MR$_{5,2}$ & MR$_{10,2}$ & \begin{tabular}[c]{@{}c@{}}Offroad\\ rate\end{tabular} & \begin{tabular}[c]{@{}c@{}}Off-yaw\\ metric\end{tabular} \\ \midrule

LVM                    & \ding{51}                     &                                                                          &                                                                                 & \ding{51}                                                                            & 1.77         & 1.27         & 0.80       & 0.63      & 0.10                                                                    & 0.12                                                                      \\
P2T$_{CS}$                  & \ding{51}                     & \ding{51}                                                                     & \ding{51}                                                                            &                                                                                 & 4.18        & 4.05         & 0.93      & 0.92        & \textbf{0.02}                                                                     & 0.07                                                                       \\
P2T$_{BC}$                 & \ding{51}                     &                                                                          & \ding{51}                                                                            & \ding{51}                                                                            & 1.47         & \textbf{1.15}         & 0.67       & 0.49      & 0.04                                                                    & 0.07                                                                      \\
P2T$_{IRL}$                & \ding{51}                     & \ding{51}                                                                     & \ding{51}                                                                            & \ding{51}                                                                            & \textbf{1.45}         & 1.16         & \textbf{0.64}       & \textbf{0.46}      & 0.03                                                                    & \textbf{0.04}                                                                      \\ \bottomrule
\end{tabularx}
\label{tab:abl_ns}
\end{table*}

\subsection{Comparison with the state of the art}
We compare our model with prior and concurrently developed models that represent the state of the art for the Stanford drone and NuScenes. 

\vspace{0.1in}

\noindent \textbf{SDD:} Table \ref{tab:sdd1} reports results on SDD based on the dataset split used in \cite{sadeghian2018sophie}. While most prior works have reported MinADE$_K$ and MinFDE$_K$ for $K$=20, Desire \cite{lee2017desire} has results reported for $K$=5. We report metrics for both values of $K$ here for our models. Note that the error values are in pixels in the bird's eye view image co-ordinates. We also report off-road rate values on SDD for our models based on per pixel path/obstacle labels for the SDD test set. Our model achieves the lowest MinFDE$_K$ values, while only being closely outperformed by the HBA-Flow model on MinADE$_K$. 

Table \ref{tab:sdd2} reports results on the dataset split used by Mangalam \textit{et al.} \cite{mangalam2020not, mangalam2020goals}. This uses a subset of trajectories in SDD, primarily consisting of pedestrians. Our model outperforms PECNet \cite{mangalam2020not}. However, the recently proposed Y-Net \cite{mangalam2020goals} achieves lower MinADE$_K$ and MinFDE$_K$ values. Similar to our models, Y-Net also conditions trajectories on goals and intermediate waypoints of agents in the scene, suggesting the importance of modeling the static scene for trajectory forecasts.  

\vspace{0.1in}
\noindent \textbf{NuScenes:} Table \ref{tab:ns} reports results on the NuScenes prediction benchmark. We compare our models with the physics oracle and CoverNet\cite{phan2020covernet} baselines released with the benchmark, and the winning entries of the NuScenes prediction challenge, cxx \cite{luo2020probabilistic}, MHA-JAM \cite{messaoud2020multi} and Trajectron++ \cite{salzmann2020trajectron++}. Additionally, we also consider the simple yet effective MTP \cite{cui2019multimodal} and Multipath \cite{chai2020multipath} models as implemented and reported on NuScenes by Greer \textit{et al.} \cite{greer2020trajectory}. Since NuScenes requires a single set of trajectories to evaluate metrics for K=5 and K=10, we merge the clustered trajectories for K=5 and K=10. To remove duplicates, we discard trajectories from the set of 10 closest to each trajectory in the set of 5 in terms of displacement error. The set of 5 trajectories is nominally assigned a higher score than the set of 10 trajectories. The benchmark does not include results for the off-yaw metric. However, we report the metric for our models as well as those from Greer \textit{et al.} \cite{greer2020trajectory}. 

Our model achieves state of the art results on almost metrics on the NuScenes benchmark\footnote{URL: \href{https://eval.ai/web/challenges/challenge-page/591/leaderboard/1659}{ https://eval.ai/web/challenges/challenge-page/591/leaderboard/1659}, results as of April 27, 2021}. In particular, it achieves significantly lower off-road rate and off-yaw metrics compared to previous methods, while still maintaining low MinADE$_K$ and miss rate values. The low MinADE$_K$ and miss rates suggest that our model generates a diverse set of trajectories. The low off-road and off-yaw metrics suggest that conditioning trajectories on plans sampled from the MaxEnt policy lead to more scene compliant trajectories. We investigate this further in section \ref{sec:ablations}.    

\subsection{Ablations}

We additionally report results for ablations of our model with respect to the following components. 

\label{sec:ablations}

\begin{figure}[t]
\centering
\includegraphics[width=\columnwidth]{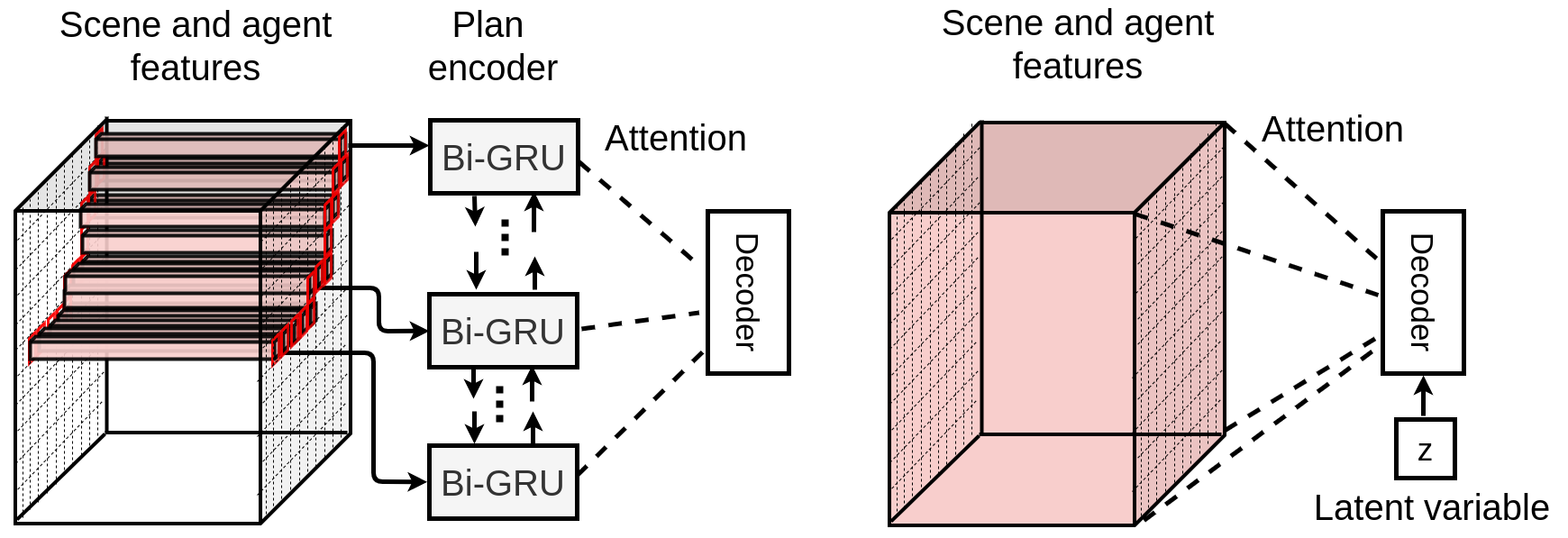}
\caption{\textbf{Ablation of grid based plans:} Models with (left) and without (right) the plan encoder and grid based policy. Without the grid based plan, the trajectory decoder attends to all features within the grid} 
\label{fig:ablation}
\end{figure}

\begin{figure*}[t]
\centering
\includegraphics[width=\textwidth]{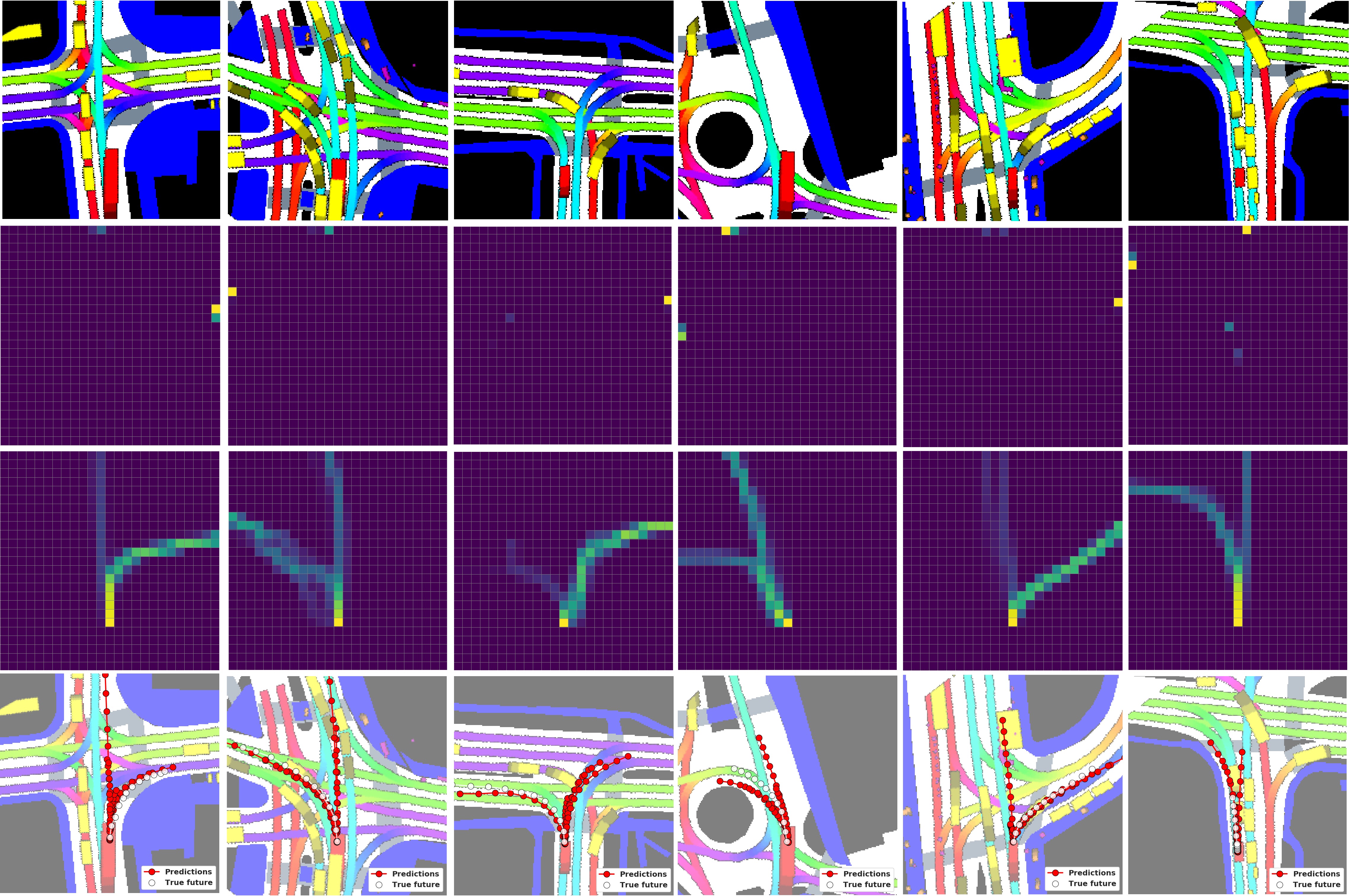}
\caption{\textbf{Qualitative examples from NuScenes}. From top to bottom: Inputs, goal SVFs, path SVFs and predictions} 
\label{fig:qual_ns}
\end{figure*}

\begin{figure*}[t]
\centering
\includegraphics[width=\textwidth]{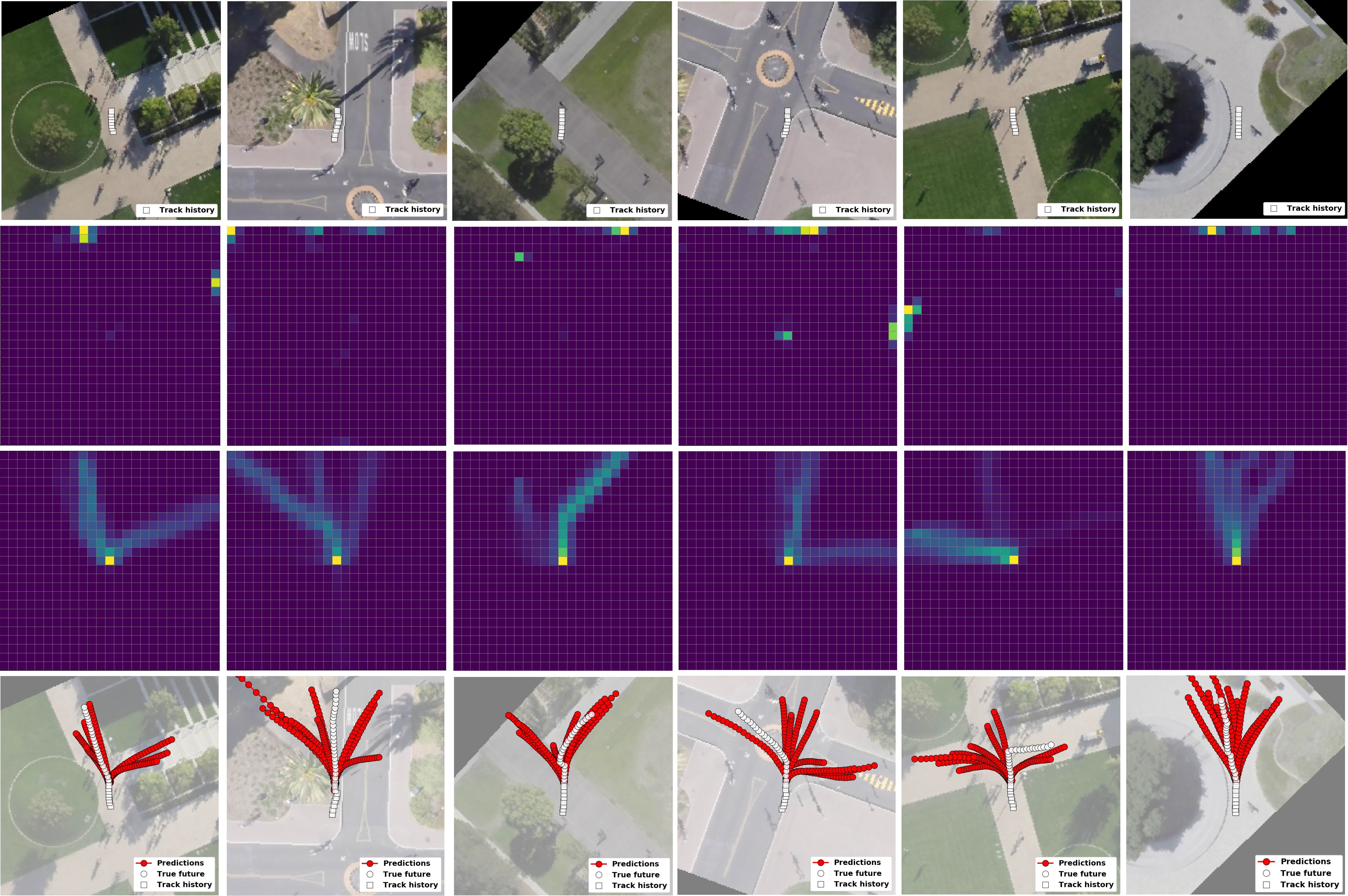}
\caption{\textbf{Qualitative examples from SDD}. From top to bottom: Inputs, goal SVFs, path SVFs and predictions} 
\label{fig:qual_sdd}
\end{figure*}

\begin{itemize}
    
\item \textbf{Grid-based plans:} To analyze the effect of conditioning trajectories on the grid based plans, we consider an ablation of our model without the MaxEnt policy and plan encoder (Figure \ref{fig:ablation}). In this case, the trajectory decoder attends to all features in the grid, rather than just those along the sampled state sequence. To keep memory usage tractable, we maxpool the features using a 2$\times$2 kernel before attention layers. In order to sample a diverse set of trajectories, we additionally condition the trajectory decoder with a latent variable $z$ sampled from a univariate Gaussian distribution. We refer to this ablation as the latent variable model (LVM). 

\vspace{0.05in}
 
\item \textbf{Trajectory generator:} Next, we consider a model without the trajectory generator. To output continuous valued trajectories along sampled plans, we fit a smoothing spline along the sampled grid locations and propagate a constant speed trajectory along the spline using the target agent's velocity at the prediction instant. We refer to this model as P2T$_{CS}$.

\vspace{0.05in}

\item \textbf{Reward Layers:} Finally, we consider an ablation without the reward layers to analyze the usefulness of using IRL. Instead of learning the reward, we learn a behavior cloning policy that directly maps the scene and motion features to action probabilities at each grid cell. We refer to this model as P2T$_{BC}$.

\end{itemize}

\vspace{0.05in}

\begin{table}[]
\caption{Inference time}
\centering
\begin{tabularx}{0.8\columnwidth}{lY}
\toprule
Component                                                                       & Time \\ \midrule
Reward model                                                                    & 2 ms     \\
Solve for MaxEnt policy (Algorithm 3)                                            & 28 ms    \\
Sample MaxEnt Policy                                                            & 28 ms    \\
Trajectory Generator                                                            & 12 ms    \\
Clustering                                                                      & 9 ms     \\ \midrule
Total                                                                           & 79 ms    \\ \bottomrule
\end{tabularx}
\label{tab:time}
\end{table}

Tables \ref{tab:abl_sdd} and \ref{tab:abl_ns} report results for ablation studies on SDD and NuScenes respectively. We note that for both datasets, our complete proposed model (P2T$_{IRL}$) outperforms the LVM across all metrics. In particular, the high off-road and off-yaw metrics for the LVM compared to the other three models suggest that the LVM generates more trajectories that veer off the road or violate lane direction. This shows that the inductive bias due to grid based plans leads to trajectories that are more scene compliant. Conversely, P2T$_{CS}$ achieves comparable offroad and off-yaw metrics as P2T$_{IRL}$. However it does poorly in terms of the MinADE, MinFDE and miss rate metrics. Thus, although its trajectories are scene compliant, they deviate significantly from the ground truth, suggesting the limitation of the constant speed model compared to the attention based GRU decoder for modeling agent dynamics. Finally, P2T$_{IRL}$ slightly outperforms the behavior cloning model P2T$_{BC}$ on most metrics, with the difference being more prominent for SDD than for NuScenes.

\subsection{Runtime}

In table \ref{tab:time} we provide inference times for each component of the model. Inference is performed using an NVIDIA GeForce GTX 1080 Ti GPU. We also implement algorithms \ref{alg:avi_goal} and \ref{alg:svf_goal} using vectorized operations on the GPU. For each prediction instance, we sample 1000 state sequences from MaxEnt policy, generating 1000 trajectories, which are finally clustered to output $K$ trajectories. The runtimes reported here are for $K$=10. We note inference can be performed in 79 ms (or ~ 12Hz) for the complete proposed model which will allow for real-time deployment, given access to rasterized birds eye view scene representations and past tracks of agents.

\subsection{Qualitative examples}
\label{sec:qual}
Figures \ref{fig:qual_ns} and \ref{fig:qual_sdd} show qualitative examples from the NuScenes and Stanford drone datasets. We show the input scene and past tracks of agents, heat maps for goal and path state visitation frequencies for the MaxEnt policy and the final set of 10 clustered trajectories from the trajectory generator. We note that the MaxEnt policy explores plausible path and goal states in the 2-D grid for a variety of scene configurations. For the Nuscenes dataset, this corresponds to reachable lanes for the target agent. Note that the policy accurately infers which lanes correspond to the direction of motion rather than oncoming traffic. For SDD, the policy shows a preference for paths and roads while avoiding terrain or obstacles. We also note that the path and goal SVFs are multimodal. Finally, the predicted trajectories closely map to the states explored by the policy, leading to a diverse set of scene compliant predictions over a variety of scene configurations.

\section{Concluding Remarks}

We introduced an approach to forecast trajectories of pedestrians and vehicles in unknown environments conditioned on plans sampled from a grid based MaxEnt IRL policy. We reformulated MaxEnt IRL to learn a policy that can jointly infer goals and paths of agents on a coarse 2-D grid defined over the scene. We showed that our policy infers plausible goals of agents in unknown environments and paths to these goals that conform to the underlying scene. Additionally, we showed that our policy induces a multi-modal distribution over path and goal states. Next, we introduced an attention based trajectory generator that outputs continuous valued trajectories conditioned on state sequences sampled from our MaxEnt policy. Trajectories sampled from our trajectory generator
are diverse and conform to the scene, outperforming prior approaches on the TrajNet benchmark split of the Stanford drone dataset and the NuScenes prediction benchmark. With an inference time of 79 ms, the proposed model can readily be deployed in conjunction with on board detectors \cite{rangesh2020ground}, trackers \cite{rangesh2019no, rangesh2021trackmpnn} and HD Maps for autonomous driving.

\ifCLASSOPTIONcaptionsoff
  \newpage
\fi



%
\bibliographystyle{IEEEtran}
\bibliography{IEEEexample}
%
\vspace{-0.4in}
\begin{IEEEbiography}[{\includegraphics[width=1.2in,height=1.25in,clip,keepaspectratio]{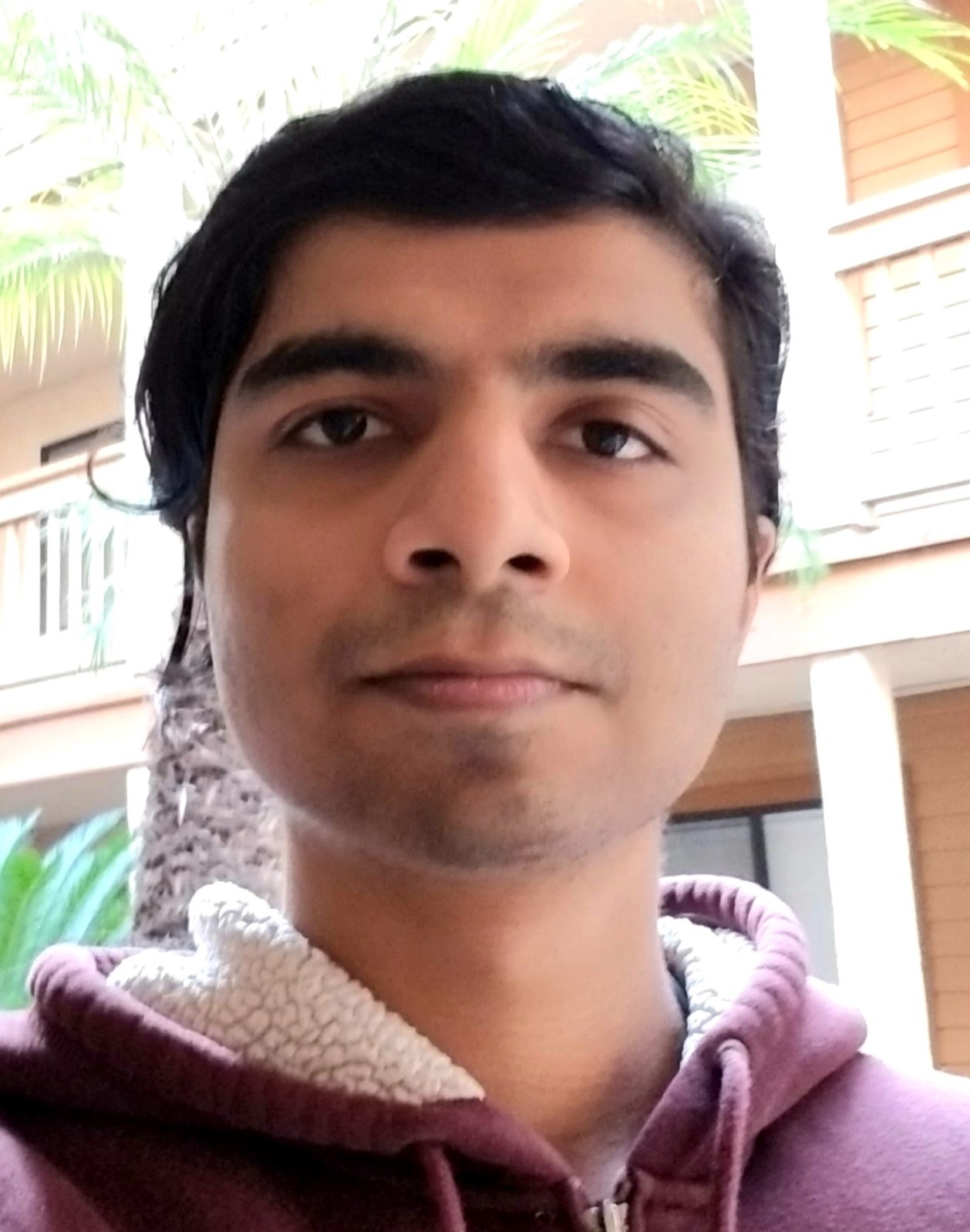}}]\\\textbf{Nachiket Deo}
is currently working towards his PhD in electrical engineering from the University of California at San Diego (UCSD), with a focus on intelligent systems, robotics, and control. His research interests span computer vision and machine learning, with a focus on motion prediction for vehicles and pedestrians.
\end{IEEEbiography}
\vspace{-0.5in}
\begin{IEEEbiography}[{\includegraphics[width=1.2in,height=1.25in,clip,keepaspectratio]{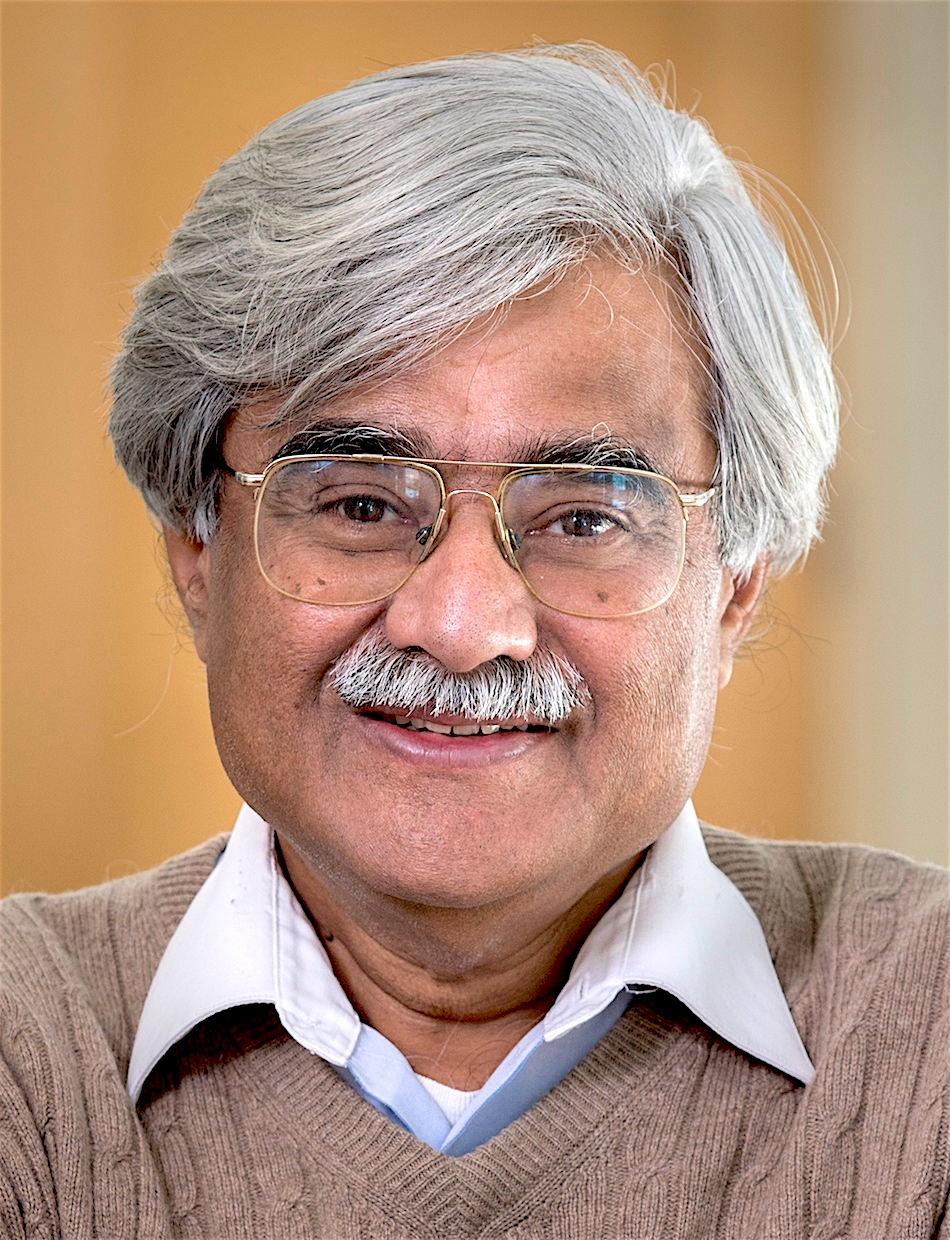}}]{Mohan Manubhai Trivedi}
is a Distinguished Professor at University of California, San Diego (UCSD) and the founding director of the UCSD LISA: Laboratory for Intelligent and Safe Automobiles,
winner of the IEEE ITSS Lead Institution Award (2015). Currently, Trivedi and his team
are pursuing research in intelligent vehicles, autonomous driving, machine perception, machine learning, human-robot interactivity, driver assistance. Three of his students have received "best dissertation" recognitions and over twenty best papers/finalist recognitions. Trivedi is a Fellow of IEEE, ICPR and SPIE. He received the IEEE ITS Society's highest accolade "Outstanding Research Award" in 2013. Trivedi serves frequently as a consultant to industry and government agencies in the USA and abroad. 
\end{IEEEbiography}
%







\end{document}